\documentclass[letterpaper, 10 pt, conference]{ieeeconf}  
\usepackage[utf8]{inputenc}
\usepackage{hyperref}
\usepackage[nocompress]{cite}
\usepackage{verbatim}
\usepackage{graphicx}
\usepackage{amsmath}
\usepackage{lipsum}
\usepackage{times}
\usepackage{breqn}
\usepackage{amsmath}

\DeclareMathOperator*{\argmin}{arg\,min}
\newcommand{\etal}{\textit{et al}.~}

\usepackage{subcaption}
\usepackage[italian]{babel}

\IEEEoverridecommandlockouts                              

\usepackage{alphalph}

\title{\LARGE \bf
Robot Calligraphy using Pseudospectral Optimal Control \\
in Conjunction with a Novel Dynamic Brush Model
}

\author{Sen Wang, Jiaqi Chen, Xuanliang Deng, Seth Hutchinson, and Frank Dellaert$^{1}$
\thanks{$^{1}$All authors are with the Georgia Institute of Technology, Atlanta, GA 30332, USA. Corresponding author:
        {\tt\small dellaert@cc.gatech.edu}}%
}

\begin{document}
\maketitle
\thispagestyle{empty}
\pagestyle{empty}

\begin{abstract}

Chinese calligraphy is a unique art form with great artistic value but difficult to master. In this paper, 
we formulate the calligraphy writing problem as a trajectory optimization problem, and propose an improved virtual brush model for simulating the real writing process. 
Our approach is inspired by pseudospectral optimal control in that we parameterize the actuator trajectory for each stroke as a Chebyshev polynomial.
The proposed dynamic virtual brush model plays a key role in formulating the objective function to be optimized. 
Our approach shows excellent performance in drawing aesthetically pleasing characters, and does so much more efficiently than previous work, opening up the possibility to achieve real-time closed-loop control. 



\end{abstract}


\section{INTRODUCTION}

Making robots write beautiful calligraphy is difficult, as learning and mastering this art form takes humans years of practice. Chinese characters are complex and a calligraphy brush is difficult to manipulate properly. Hence, making a robot achieve comparable results is a worthwhile endeavour, both to expand robot's capabilities into art as well as push our understanding on how to manipulate deformable brushes. 

Most relevant research in this area adopts either a learning-based method or an approach based on trajectory optimization.
The former includes learning from demonstration~\cite{Sun14iros_robot-learn-from-demo, Kotani19icra_TeachingRT}, or learning from visual feedback given a reference image~\cite{Mueller13iros_robotic_calligraphy}. 
By using learning, one can eliminate the difficulty of modeling the behavior of a real calligraphy brush. However, learning methods have a large training cost and may not generalize well to previously unseen characters.
On the other hand, trajectory optimization methods do not face these problems: they \emph{simulate} the writing behavior of an actual brush, and then search for an optimal trajectory for the robot to execute~\cite{Kwok06icec_Brush_stroke_generation,Lam09iros_Stroke_Trajectory}).
The difficulty there is that most simulated brush models~\cite{Kwok06icase_robot_drawing,Lam09iros_Stroke_Trajectory} do not account for the complex ways a brush deforms during the writing process. Being able to capture the complexity of a deformable brush has an important influence on the final performance. 

\begin{figure}[htb]
    \centering
    \begin{subfigure}[b]{0.23\textwidth} 
        \centering
        \includegraphics[width=0.7\linewidth]{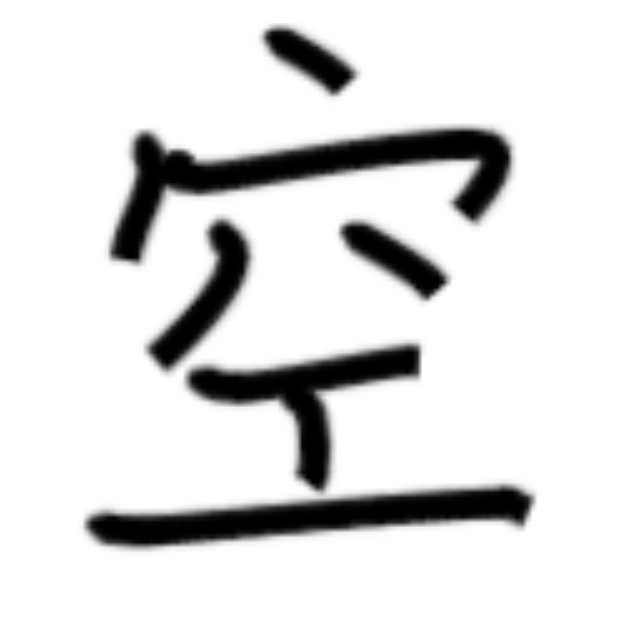}
        \caption{Simulated image from initial trajectory estimate}
    \end{subfigure}
    \begin{subfigure}[b]{0.23\textwidth}
        \centering
        \includegraphics[height=0.7\linewidth]{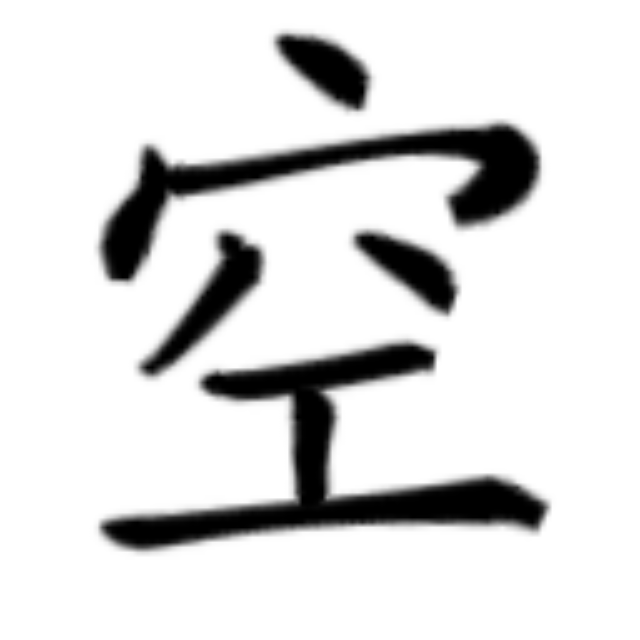}
    \caption{Simulated image after trajectory optimization}
    \end{subfigure}
    \vskip\baselineskip
    \begin{subfigure}[b]{0.23\textwidth}
        \centering
        \includegraphics[width=1.0\linewidth]{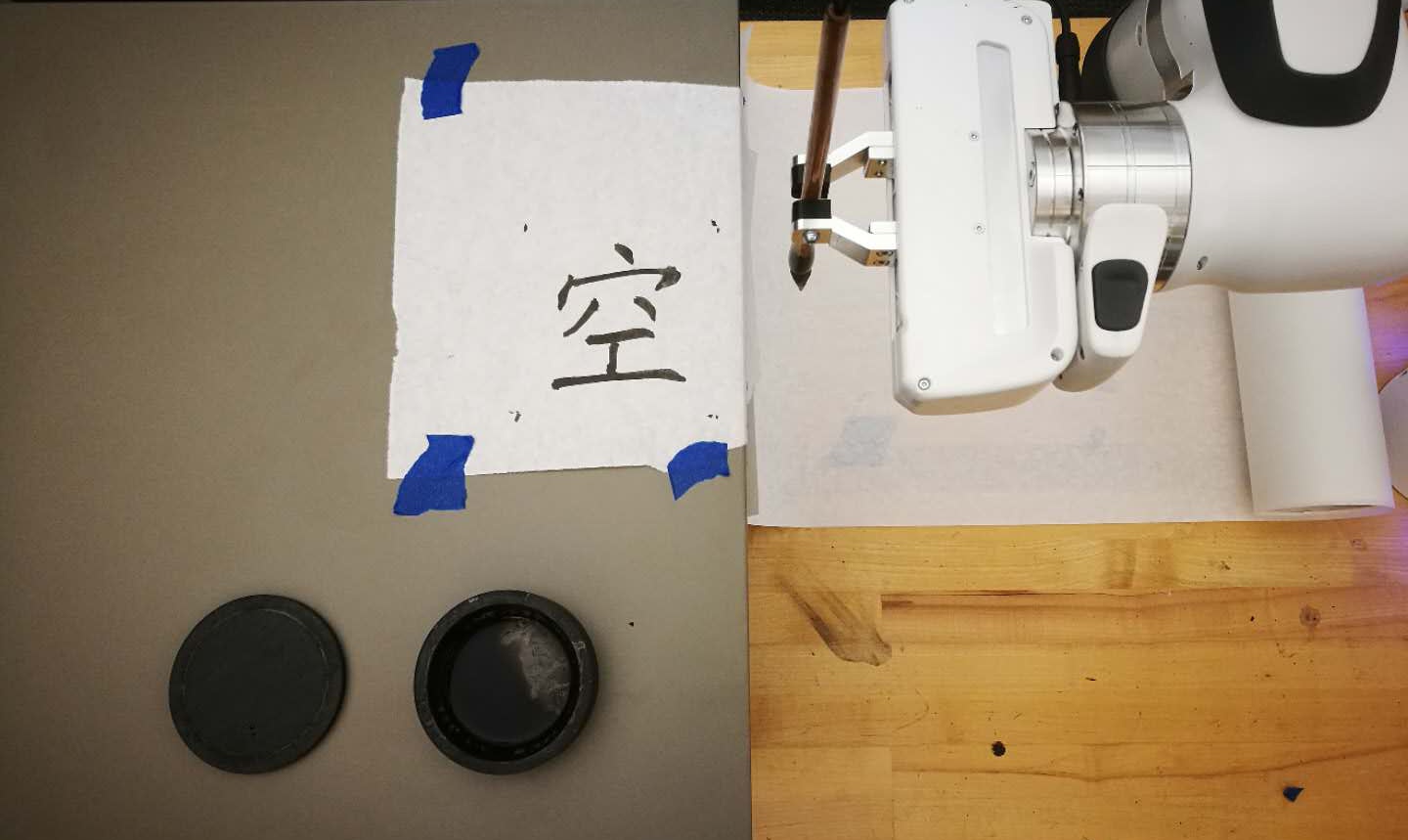}
        \caption{Robot executing trajectory}
    \end{subfigure}
    \begin{subfigure}[b]{0.23\textwidth}
        \centering
        \includegraphics[height=0.7\linewidth]{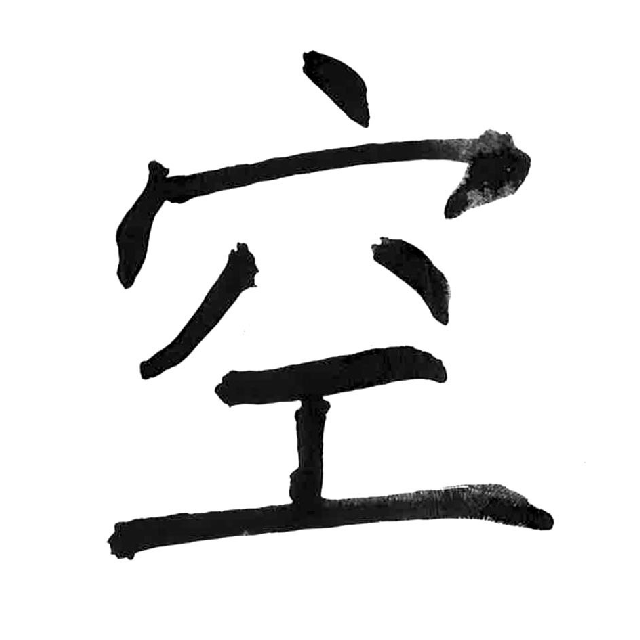}
    \caption{Written image}
    \end{subfigure}
    \caption{Summary of project workflow}
    \label{four process}
\end{figure}

In this paper, we propose a novel virtual brush model intended to capture the dynamics of an actual calligraphy brush, which is an improved version based on an earlier virtual brush model by Kwok \etal\cite{Kwok06icec_Brush_stroke_generation}, which we use as a baseline. We improve on this model by explicitly modeling the brush dynamics, while retaining a much simpler structure compared to the elaborate models in other work~\cite{Xu09book_Calligraphy,Chu04icga_realtime_virtualBrush}. We then use this model to simulate the writing process for a given open-loop trajectory, which allows us to then optimize for the trajectory parameters.
%

To do so, our second contribution is an efficient trajectory optimization-based method based on \textit{pseudospectral optimal control}~\cite{Elnagar98coa_Chebyshev,Fahroo00acc_DirectTO} that achieves fully automatic writing of Chinese characters given a character's unicode. 
Pseudospectral methods are based on Legendre or Chebyshev polynomials, which are excellent at approximating and representing continuous trajectories and controls. Our method converges quickly and efficiently even when using many control points.


Finally, we exploit the existence of vector-based character databases such that, given a character's unicode, we immediately have access to the individual character strokes as well as the stroke order. 
We optimize for each stroke separately, and to obtain even faster convergence we initialize the trajectory for each stroke by a stroke skeleton which is also available in these databases.

Our proposed continuous, nonlinear optimization framework differs from previous work~\cite{Kwok06icec_Brush_stroke_generation, Kwok06icase_robot_drawing, Lam09iros_Stroke_Trajectory} which uses heuristic optimization methods known for slow convergence rates and a high computational cost. Hence, our method can potentially be used in a closed-loop control system, which we intend to pursue in future work.
The proposed end-to-end system displays excellent writing performance, although there is definitely room for growth.

\section{Related Work}
There is much work in the robotics community that aims to create art using robots, such as painting and drawing \cite{Kudoh07iros_painter-robot,Lu09icaim_PreliminarySO,Scalera18jirs_watercolor}, sculpture\cite{Niu07robio_Robot3S}, graffiti~\cite{Jun16iros_Humanoid_Graffiti}, etc. 
Below we focus our discussion on robot calligraphy, most algorithms using a calligraphy brush can be categorized as either learning or trajectory optimization-based.

\subsection{Calligraphy robots using learning-based methods}
Examples of simple learning-based methods include Sun \etal's learning from demonstration ~\cite{Sun13robio_callibot}
\cite{Sun14iros_robot-learn-from-demo}, Mueller \etal's trial-and-learn iterative learning method ~\cite{Mueller13iros_robotic_calligraphy}. 
Some more advanced learning algorithms such as RNN~\cite{Sasaki16ras_visual_motor}, generative adversarial networks~\cite{Chao18icra_calligraphy_gan}, deep reinforcement learning~\cite{Wu18robio_DRL_calligraphy}, and local and global learning models~\cite{Kotani19icra_TeachingRT} were also explored.
However, these methods require many iterations of training to achieve good performance, and have difficulty generalizing to new and/or complicated characters.


\subsection{Virtual brush models}
Trajectory optimization-based algorithms mainly rely on \textit{virtual brush models}, which can be divided into two categories: physics-based models and data-driven models.

Physics-based virtual brush models strive to simulate the physical dynamics of a real brush from experimental observation~\cite{Xu02cgf_hairy_brush, Xu03_CGF_virtual_brush, Saito93book_DPB,Lee99book_BSO}  or physical laws~\cite{Chu02pccga_3dpainting, Chu04icga_realtime_virtualBrush}. Strassmann~\cite{Strassmann86siggraph_hairy-brush} proposes an initial design featuring four basic parameters of a hairy brush based on bristles. 
Wong \etal~\cite{Wong00cg_virtual_brush} propose to use a cone to represent the bundle of the brush and use the cross-section of the cone, an ellipse, to represent the footprint of the brush. 
Xu \etal~\cite{Xu09book_Calligraphy} propose a detailed virtual brush model for use in synthetic imagery, with good results in creating realistic-looking simulations. 
However, obtaining and fitting good parameters to the models above is difficult. As such, we propose a virtual brush with an easy structure to fit and implement, geared towards real-time trajectory optimization.

 
Data-driven virtual brush models are created by measuring and recording actual brush footprints on writing surfaces. Kwok \etal  propose a simple virtual brush which draws droplet shapes whose size is proportional to the writing height \cite{Kwok06icec_Brush_stroke_generation}. In their later work~\cite{Kwok06icase_robot_drawing}, they use a camera placed below the writing plane to collect footprints during the writing process. 
Lam \etal~\cite{Lam09iros_Stroke_Trajectory} define their writing mark as a polygon connected by eight points and fit their position parameters with the collected footprints. 
With an eye for efficiency, Baxter \etal~\cite{Baxter10book_SimpleDM} build a deformation table to speed up the associated computation while still being able to simulate complex effects. 



\subsection{Stroke extraction}


Stroke extraction involves separating a character into its comprising strokes and is difficult to do with good accuracy when analyzing only the pixels of an image. There are three main categories of stroke extraction methods: skeleton-based~\cite{Liu01pr_ModelbasedSE,Zeng10is_cmrf_stroke}, region-based~\cite{Cao00icpr_stroke_extraction}, and contour-based~\cite{Lee98pr_stroke_extraction,Sun14icra_brush-stroke-extraction,Chao19_HMS_calli_style}. 
Most of these methods are complex and may not achieve good results as compared to the ground-truth stroke segmentation. As such, we propose using vector-based character database, which provides a quick and accurate way to extract strokes, as well as stroke order.




\subsection{Optimization methods}
As mentioned above, Kwok \etal~\cite{Kwok06icec_Brush_stroke_generation,Kwok06icase_robot_drawing} propose using genetic algorithm for direct character-level optimization guided by their brush models. However, their reported computation times are long, and the strokes have to be  manually separated. 
Lam \etal~\cite{Lam09iros_Stroke_Trajectory} minimize the width difference of the strokes between reference images and a simulated image written by the virtual brush as it moves along the middle axis of each stroke. However, their method is sensitive to small variations in stroke images, and so the results suffer from a loss of smoothness. One of our main contributions below is the use of methods from optimal control to achieve the desired stroke trajectories.




\section{Methodology Overview}
\begin{figure*}[ht!]
  \centering
  \includegraphics[width=\textwidth]{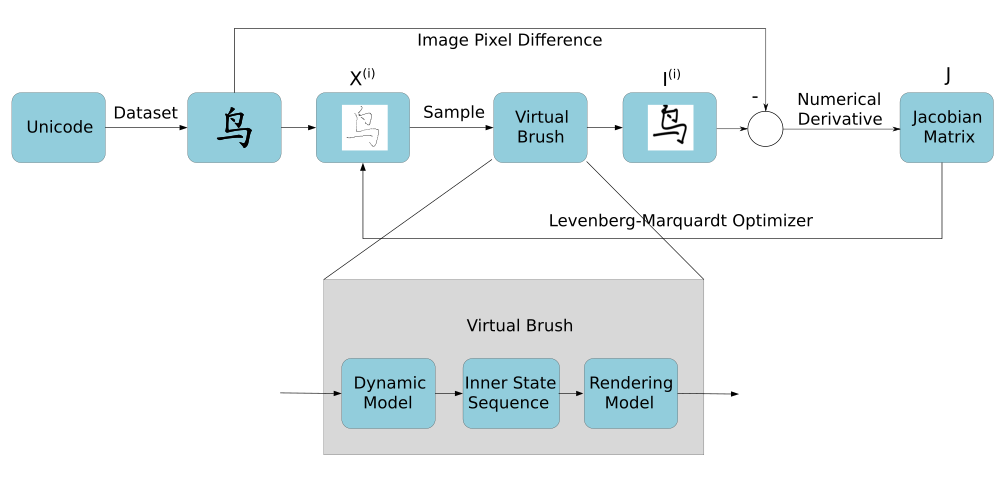}
  \caption{Flowchart of the Methodology. (1) Given a unicode, find the corresponding image in the database; (2) Extract the initial trajectories from the character image; (3) Use the virtual brush to draw a simulated image, and then compute the numerical derivative from the previously drawn image if it exists; (4) Use the Jacobian matrix to improve the initial trajectory; (5) Continue the loop until the stopping criteria are satisfied.}
  \label{flowchart}
\end{figure*}

An overview of our methodology is given in Fig.~\ref{flowchart} on the next page. 
Given a character's unicode we find the corresponding SVG image in a vector-based database, which already comes segmented into the different strokes. For each stroke, we compute initial trajectories from the stroke skeletons, and then optimize from there until our simulated image matches the reference images rendered from the SVG file. At the core of the simulation is a virtual brush model, which we discuss at length in the section below, followed by a discussion of our optimization framework.

\section{Two Virtual Brush Models}
There are two virtual brush models that we have experimented with in order to simulate the calligraphy writing process: a simple virtual brush and a dynamic virtual brush. Both generate good results, but the dynamic virtual brush is more sophisticated and better at closing the ``Sim2Real" gap. 


\subsection{Simple virtual brush model}

The simple virtual brush model is similar to Kwok \etal's work ~\cite{Kwok06icec_Brush_stroke_generation} and is used as a baseline comparison in our results.  
Given the height $z$ of the brush, this model simply fills a circular patch with simulated ink, where the radius of the circle is proportional to $z$.
Note that we align the task coordinate frame for the robot and brush such that the $z=0$ plane corresponds to the plane of the paper and z increases as the brush is moved towards the paper.
Note that Kwok \etal also proposed using more complex templates (such as droplets), but we get that behavior for free by modeling the dynamics in our model below.

\subsection{Dynamic virtual brush model}
\newcommand{\state}{\mathcal{X}}
\newcommand{\control}{\mathcal{U}}
\newcommand{\KK}{K}
\newcommand{\PP}{\boldsymbol{X}}
\newcommand{\Root}{\PP^{root}}
\newcommand{\VV}{\mathcal{V}}

The dynamic virtual brush model has two components: a drawing component, and a dynamic update component. The drawing component describes how the brush leaves a mark on paper depending on its parameters.  The updating component then describes how the brush parameters are updated due to deformations when executing an open loop trajectory $[x(t),y(t),z(t)]^T$. 

\subsubsection{Drawing Component}

\begin{figure}
\centering
\includegraphics[width=6.5cm]{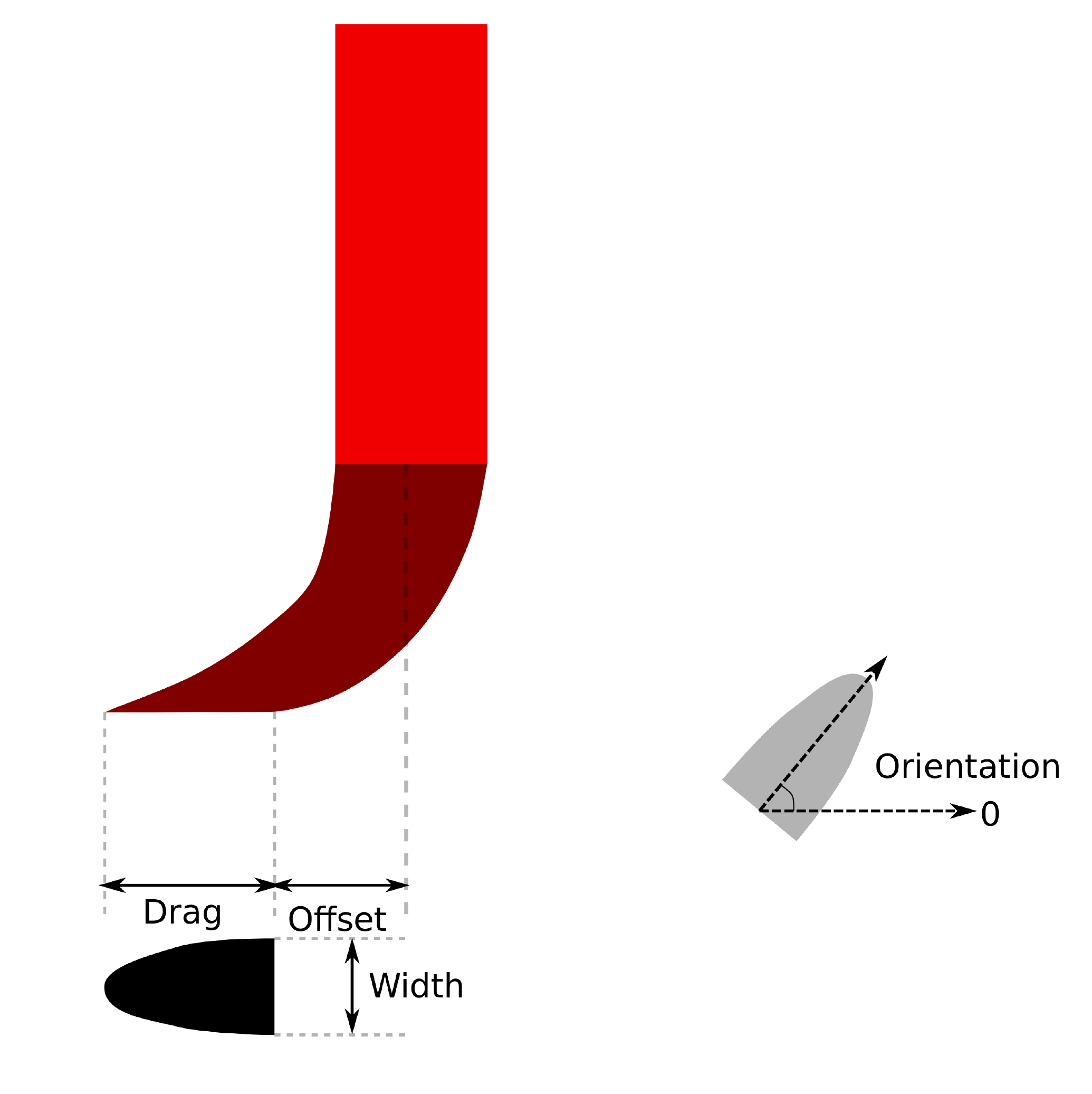}
\caption{Dynamic virtual brush model and its parameters. The root location of the brush mark is defined as the middle of the flat end and is determined by the end-effector position $\PP$, the offset $o$, and the orientation $\theta$, whereas the shape of the brush mark is determined by the width $w$ and drag $d$.}
\label{virtual_brush}
\end{figure}

The dynamic virtual brush model has four state parameters governing its drawing footprint: width $w$, drag $d$, offset $o$, and orientation $\theta$. As shown in Fig.~\ref{virtual_brush}, width $w$ and drag $d$ define the size of a brush mark, the offset $o$ simulates the deviation of the brush mark from the center of the vertical brush handle due to bending of the brush hairs when applying pressure, and the orientation $\theta$ describes the direction of the writing mark. 
Together with the end-effector coordinates $\PP \doteq (x,y,z)$, these four parameters define the 7-dimensional brush state $\state \doteq (x,y,z,w,d,o,\theta)$.

Given the state $\state$, our model then computes the ``root" location of the brush mark as 
\begin{align}
\Root = \pi(\PP) - o \VV (\theta) \label{eq:root}
\end{align}
where $\pi(\PP)$ is the projection of $\PP$ in the $z=0$ plane and $\VV$ transforms the scalar orientation $\theta$ into a three-dimensional unit vector in that same plane.
The shape of the brush mark is parameterized by a quadratic curve which is uniquely determined by the width $w$ and the drag $d$. Some more specialized brush behavior such as intentional hair-splitting is reserved for more artistic calligraphy styles and is not modeled here.


\subsubsection{Dynamic Update Component}
\begin{figure*}[ht]
\centering
\includegraphics[width=6in]{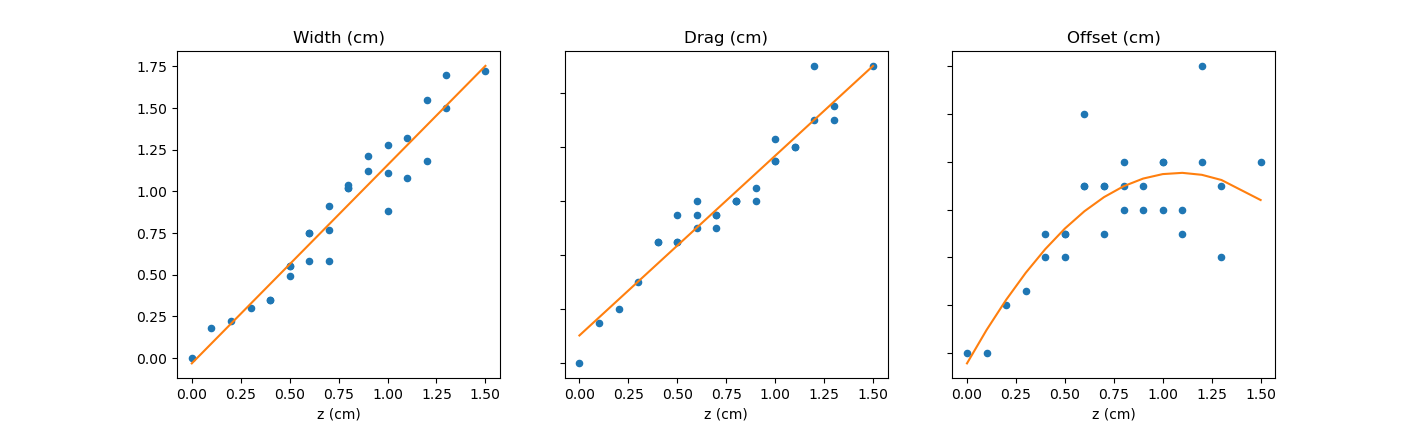}
\caption{The data collected for fitting the model parameters for our new virtual brush model (2 outliers were removed).}
\label{data_fitting_vb}
\end{figure*}
We adopt a simple discrete state dynamical model described at a high level by
\begin{align}
\state_{i+1} = f(\state_i, \control_i) \label{eq:dynamics}
\end{align}
where $\control_i=(v_{xi}, v_{yi}, v_{zi})$ is the control at the $i^{th}$ time instance.
The $x$, $y$, and $z$ components of the state $\state_0$ at $i=0$ are initialized from the first point in the open loop trajectory. 
Initializing the other components is more involved and is discussed in more detail in Section \ref{sec:trajopt}.

The dynamics model is given below and was formulated based on careful observation and experimentation:
\begin{align}
    x_{i+1} &= x_i + v_{xi} \Delta t\\
    y_{i+1} &= y_i + v_{yi} \Delta t\\
    z_{i+1} &= z_i + v_{zi} \Delta t\\
    w_{i+1} &= w_{i} \KK_w + \text{\textit{Width}}(z_{i+1})(1 -  \KK_w) \label{eq:width} \\
    d_{i+1} &= d_{i}\KK_d + \text{\textit{Drag}}(z_{i+1})(1 -  \KK_d) \\
    o_{i+1} &= \min(\textit{Offset}(z_{i+1}), \ \textit{Offset'}(\Root_i,  \PP_{i+1})) \\
    \theta_{i+1} &= \VV^{-1}(\Root_{i+1} - \pi(\PP_{i+1})) \label{eq:theta}
\end{align}
Above \textit{Width}$(z_{i+1})$ and \textit{Drag}$(z_{i+1})$ respectively compute a new value for width and drag, given the updated end-effector height $z_{i+1}$.  The constants $\KK_w$ and $\KK_d$ model a small amount of inertia in those parameters - we have used 0.02 in our experiments - modeling the fact that the deformations happen gradually and steadily.

The orientation and offset updates are more complicated given that they are governed by friction, modeling the fact that the part of the brush in contact with the paper resists movement. In particular, the function \textit{Offset}$(z_{i+1})$ computes a value for the offset $o$ when friction is overcome, and \textit{Offset'}$(\Root_i, \PP_{i+1})$ computes the value of $o$ that makes the brush stay put. The $min$ operator ensures that the brush stays put until it ``snaps" to the smaller value \textit{Offset}$(z_{i+1})$ when the end-effector $\PP_{i+1}$ moves too far out. The orientation $\theta_{i+1}$ is accordingly updated, as $\Root_{i+1}$ is a function of $o_{i+1}$.


\subsubsection{System Identification}

The width, drag, and offset functions above were experimentally determined in a separate calibration phase.
We measured the width, drag, and offset of writing marks left by a real calligraphy brush, then fit a polynomial relationship between the parameters, width $w$, drag $d$, and offset $o$, and the change in height of the brush, $z$. 
We fit the relationships between the parameters width $w$, drag $d$, and offset $d$ w.r.t. $z$ by collecting brush footprint data at varying writing heights, shown in Figure \ref{data_fitting_vb}.
The figure shows that width and drag behave almost linearly, but that the offset $o$ has a local maximum.
We observe that moving the brush in one direction will cause the tip of the brush to gradually move towards the opposite direction.

\section{Trajectory Optimization}
\label{sec:trajopt}

\subsection{Review of Pseudospectral Optimal Control}

Below we formulate the calligraphy writing process as a trajectory optimization problem, and in particular we adopt some of the machinery from pseudospectral optimal control (PSOC) methods. As such, we briefly review pseudospectral optimal control methods in this section, closely following the exposition by Fahroo \etal~\cite{Fahroo00acc_DirectTO}.

A simplified version of an optimal control problem can be stated in terms of a cost function $C$ and system dynamics $f$
    \begin{align}
        C = g(x, u) \label{eq: obj}\\
        \dot{x}(t) = f (x(t), u(t), t)),
        \label{eq: dyna}
    \end{align}
where the objective is to find the optimal control sequence $u(t)$ that minimizes the cost function $C$. Above, $x(t)$ represents the system's state trajectory.

The basic idea in PSOC is to approximate the control trajectory $u(t)$ \textit{and} the state trajectory $x(t)$ by a polynomial curve with unknown parameters, thereby transforming the original problem into a nonlinear programming problem. To this end, pseudospectral methods choose a specific set of points from the curve for interpolation. For example, in the case of Chebyshev pseudospectral methods, 
the interpolation points used are the Chebyshev-Gauss-Lobatto (CGL) points~\cite{Furgale12icra_ContinuoustimeBE}:
\begin{equation} \label{eq:cgl}
\begin{split}
t_k = \cos \left( \frac{\pi k}{N} \right) , \ \ \ \ k=0, ..., N 
\end{split}
\end{equation}
To recover the state $x(t)$ at any arbitrary time $t$ one can use barycentric interpolation,
i.e.
\begin{equation}\label{eq: x_approx}
\begin{split}
x(t) \approx (\sum_{k} \frac{w_k}{t-t_k}x_k )/ \sum_{k} \frac{w_k}{t-t_k}
\end{split}
\end{equation}
where
\begin{equation}\label{eq:w}
  w_k =
  \begin{cases}
   (-1)^k/2 & \text{$k=0$ or $k=N$} \\
  (-1)^k & \text{otherwise}
  \end{cases}
\end{equation}

We can now express the original dynamics equation with an approximation, where the objective function ~\eqref{eq: obj} can be discretized if necessary. The original problem is thus transformed to minimizing the cost $C$ with respect to the two coefficient vectors $X,\ U$:
\begin{equation} 
\begin{split}
X=(x_0,...,x_N),\ U=(u_0,...,u_N)
\end{split}
\end{equation}
representing the values of the state and controls, respectively, at the CGL points.

\subsection{PSOC for Calligraphy}

Below we apply these methods to trajectory optimization for open-loop control trajectories of a robot end-effector, with the goal of faithfully reproducing Chinese characters. PSOC methods are generally used for collocated optimal control where the system dynamics are enforced through specialized components of the cost function. However, because our control input is simply the trajectory velocity, which can be calculated exactly from the end-effector trajectory, we have no need for separate control parameters $U$.

The optimization for a character is decomposed into a series of trajectory optimization problems corresponding to individual strokes of the character. This greatly simplifies the process and is also more computationally efficient. In this, we are helped by the existence of vector-based character databases in which characters are stored as their decomposed individual strokes, as described next.

\subsection{Stroke extraction}

\begin{figure}[h]
\centering
\includegraphics[width=2.5in]{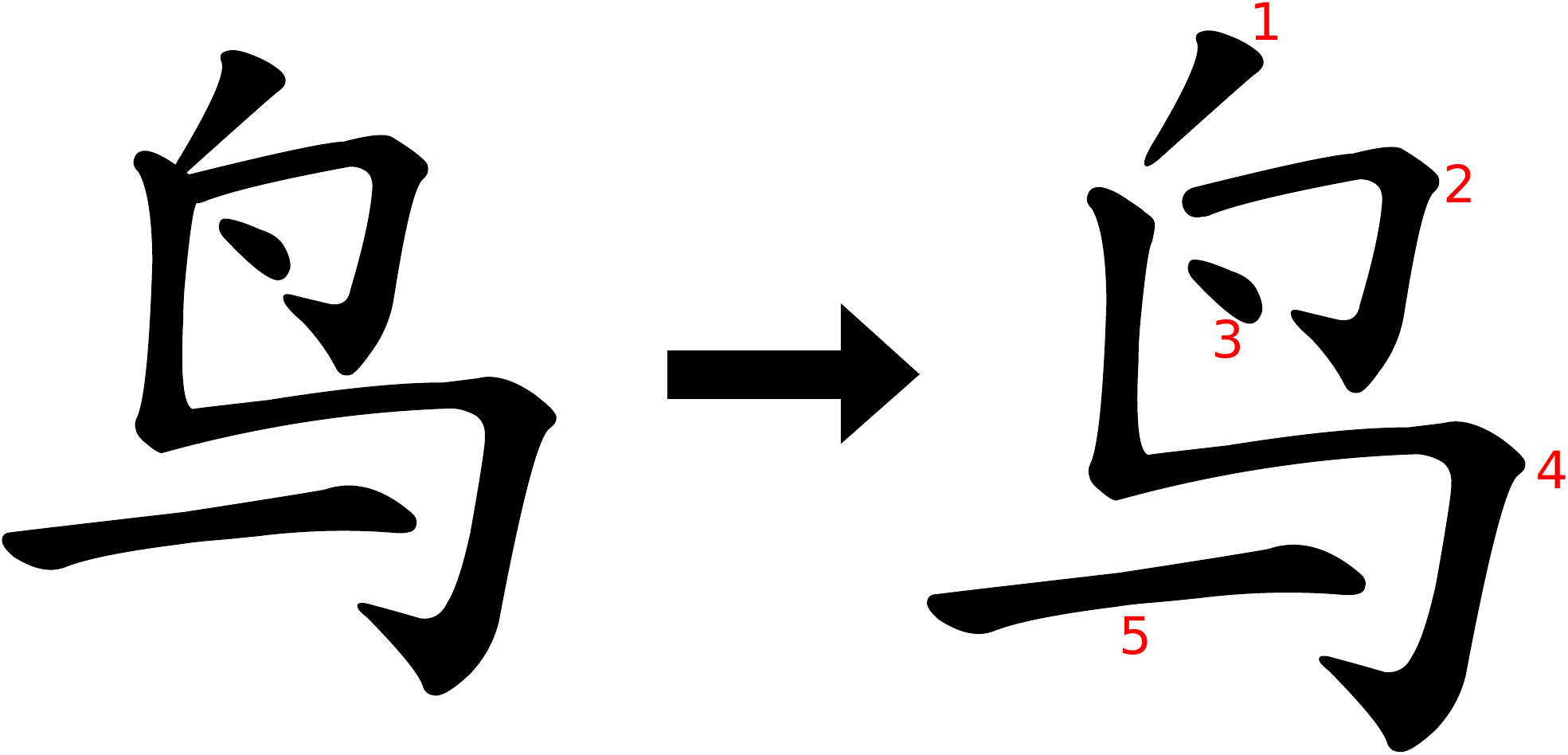}
\caption{The character 'bird', pronounced 'niao', and its extracted strokes}
\label{stroke_extraction}
\end{figure}

To obtain reference images for each stroke, as well as initialize the nonlinear optimization, we exploit the existence of vector-based character databases.
Vector-based images are advantageous because they store individual strokes as Bezier curves, which are themselves continuous polynomial curves defined by a set of control points. This makes the stroke extraction a trivial task. In our case, the database we choose is a Scalable Vector Graphics (SVG) database from \href{https://github.com/skishore/makemeahanzi}{MakeMeHanzi} \cite{makeahanzi}, and an example of extracted strokes can be seen in Fig.~\ref{stroke_extraction}.

\subsection{Stroke trajectory representation}

The stroke trajectories will be used as the open-loop control trajectories for the robot to draw strokes, and we represent them as three-dimensional trajectories of the end-effector. Each of the $x, y, z$ components are separately represented as 1-dimensional Chebyshev polynomial curves.
The trajectories $\PP$ with three dimensions $x,y,z$ are expanded by interpolating the values at the Chebyshev-Gauss-Lobatto (CGL) points given by Eq.~\ref{eq: x_approx} and Eq.~\ref{eq:w}:
\begin{equation} 
\begin{split}
\PP(t) \approx (\sum_{k} \frac{w_k}{t-t_k}\PP_k) / \sum_{k} \frac{w_k}{t-t_k}
\end{split}
\end{equation}
Hence, the decision variables are the combination of the three sets of CGL points in the $x,y,z$ dimensions:
\begin{equation} \label{eq:x_diff}
\begin{split}
X=(x_0,...,x_N ; y_0,...,y_N ; z_0,...,z_N)
\end{split}
\end{equation}


\subsection{Optimization for the stroke trajectories} 
The objective function $C_s$ for stroke $s$ minimizes the sum-squared pixel difference between the image $V(X_s)$ produced by simulating the drawing process, and a reference image $I_s$ created from the SVG representation. The optimal trajectory parameters $X_s^*$ for stroke $s$ are obtained as
\begin{align}
\label{eq:regularized}
X_s^* &= \argmin_{X_s} C_s(X_s) \\
&= \argmin_{X_s} \{ \| \mathit{V}(X_s) - \mathit{I_s} \|^2 + \sum_k \beta_k \|z_k\|^2 
\}
\end{align}
Above, $\mathit{V}(\cdot)$ is a function that represents the virtual brush simulation, taking a stroke trajectory with pseudospectral parameterization $X_s$ and drawing a stroke image according to the given trajectory. 
The $\beta_k \|z_k\|^2$ are regularization terms on the brush height to make sure the trajectory has a tendency to lift up the brush near the end of the stroke. The weights $\beta_k$ were determined by experiment. 

The optimization is conditioned on the initial brush state $\state_{init}$. The position is initialized from the trajectory, whereas the width $w$, drag $d$, offset $o$, and orientation $\theta$ are simply set from the final values of the previous brush stroke.

In practice we try all polynomial orders between 3 and 8 for each stroke and pick the one with the smallest error, in order to adapt to strokes of varying length and complexity.
A dense and even sampling of the polynomial is performed in $V$ to execute a continuous stroke in simulation.

We use the Levenberg-Marquardt (LM) algorithm, a second-order trust-region method, to minimize Eq.~\ref{eq:regularized}. In short, LM switches between a gradient-based search and a second-order Gauss-Newton update by controlling a \textit{damping factor} $\lambda$:
\begin{equation}
     [\mathbf{J}^T \mathbf{J} + \lambda \textbf{diag}(\mathbf{J}^T \mathbf{J}) ]\delta = \mathbf{J}^T[\mathit{V}(\mathit{c_i}) - \mathit{I_i}] 
\end{equation}
The Jacobian matrix $\mathbf{J}$ is computed using numerical differentiation at each iteration. We use the the GTSAM library~\cite{Dellaert12book_FactorGA} to perform the optimization. GTSAM was originally created to solve simultaneous localization and mapping problems but has been used in many different contexts since, including motion planning~\cite{Mukadam18ijrr_gpmp, Mukadam18ar_steap}.

\subsection{Trajectory Initialization Estimates}
\label{sec:init}
We use the skeleton of a stroke to initialize the $x$ and $y$ coordinates, while the initial $z$ coordinates are set to a fixed value. From our observations, when people write calligraphy, they generally make the brush approximately follow the skeleton of the stroke while varying the height of the brush. 

\begin{figure}[h]
    \centering
    \begin{subfigure}[b]{0.15\textwidth}
        \centering
        \includegraphics[width=\linewidth]{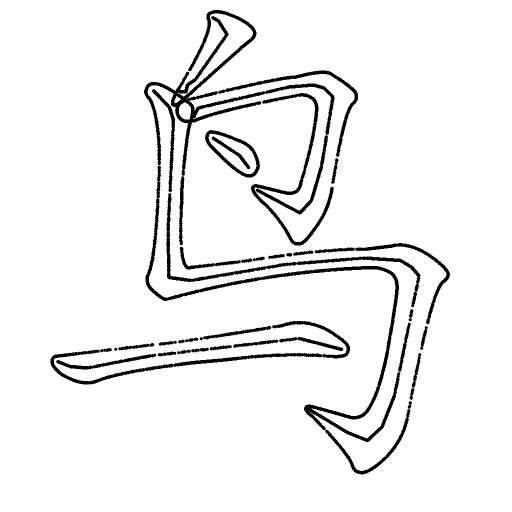}
        \caption{}
        \label{bird_svg}
    \end{subfigure}
    \begin{subfigure}[b]{0.15\textwidth}
        \centering
        \includegraphics[height=\linewidth]{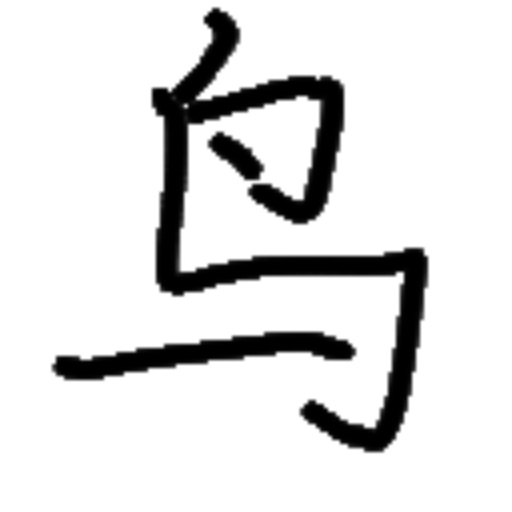}
        \caption{}
        \label{initial_simple}
    \end{subfigure}
    \begin{subfigure}[b]{0.15\textwidth}
        \centering
        \includegraphics[height=\linewidth]{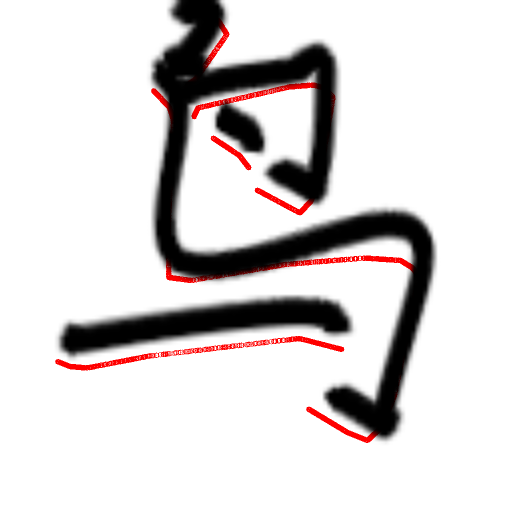}
        \caption{}
        \label{initial_dynamic}
    \end{subfigure}
    \caption{(a) The character `bird' with its skeleton; (b) Initial image from simple virtual brush; (c) Initial image from dynamic virtual brush with the trajectory (red) }
\end{figure}

Because we start from a vector-based representation, extracting an initial trajectory for the individual strokes is much simplified. Even when starting from images, there are many good image-based skeleton extraction algorithms, e.g. the Chordal Axis Transform~\cite{Zhang96icpr_thinning,Lo06iros_BrushFA}, an example result of which is shown in Fig.~\ref{bird_svg}. In our case, we use the ``animation path" provided by the database as the initial estimation for the 2D $x_i$ and $y_i$ sequence for simplicity. CGL points are sampled on each skeleton path to obtain the pseudospectral representation.
Generating an easy estimate for $z$ is not intuitive, and so we just initialize it with a constant sequence. 

Given the initial trajectory, we can simulate the image formation process using both the simple virtual brush and the dynamic virtual brush, as illustrated in Fig.~\ref{initial_simple} and Fig.~\ref{initial_dynamic} respectively. The position of the written mark from the dynamic virtual brush is different from its given trajectory, which is ignored by most previous research.

\subsection{Improving the Sim2Real Gap}

Part of the Sim2Real gap is due to the unpredictable state of the brush after executing a stroke.
After the brush writes a complicated stroke the tip may become messy. 
At that time, accurately predicting the state of the brush and finding a feasible control trajectory is difficult and unreliable. 
Writing brushes also need to periodically dip ink to be able to write. 

To avoid accumulated prediction error as more strokes are written, we have the robot \textit{dip ink} after a stroke is written, which restores the brush to a predictable state.
We handcrafted a control algorithm to accomplish this:
given a circular ink stone, the brush is pushed down heavily at first to make the tip flat, and we then slowly move it to the edge of the ink stone in different directions with gradually smaller extent. 
After that, the tip is generally restored to a predictable state, which we reflect in the initial state $\state_0$ of the dynamic brush model.
In particular, width $w$, drag $d$, offset $o$, are set to zero, whereas the orientation $\theta$ is initialized to align with initial trajectory direction.

\section{Results}

\begin{figure}[h]
    \centering
    \begin{subfigure}[b]{0.15\textwidth}
        \centering
        \includegraphics[width=\linewidth]{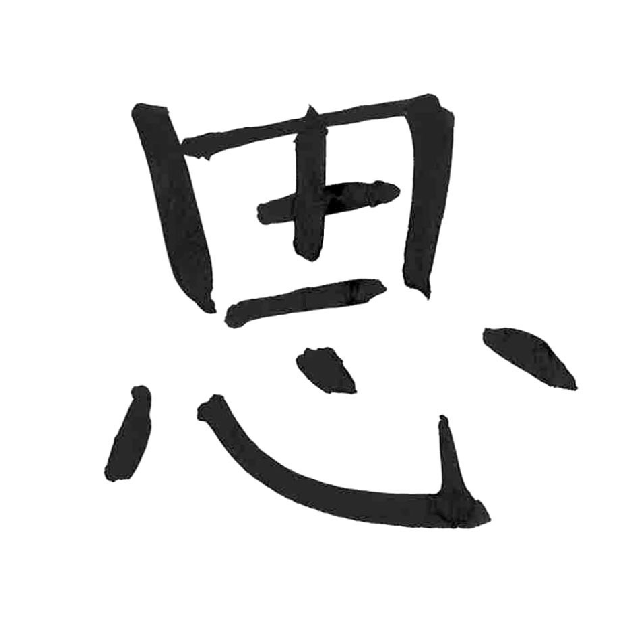}
        \caption{}
        \label{bird_svg2}
    \end{subfigure}
    \begin{subfigure}[b]{0.15\textwidth}
        \centering
        \includegraphics[height=\linewidth]{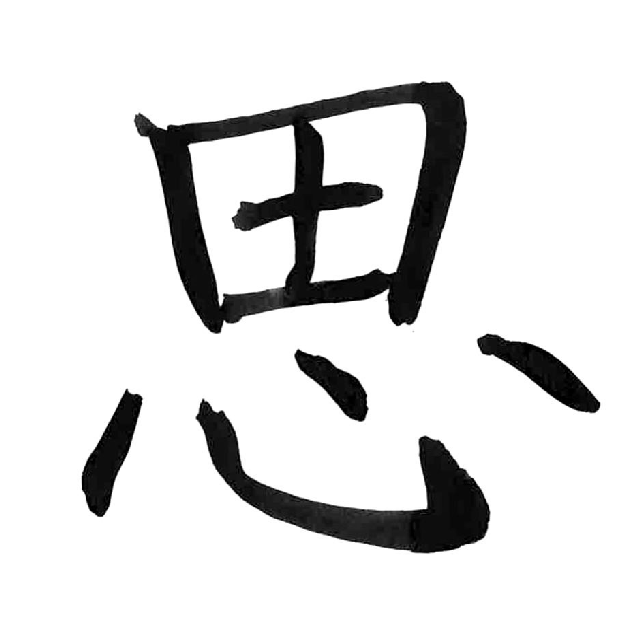}
        \caption{}
        \label{initial_simple2}
    \end{subfigure} \\
    \begin{subfigure}[b]{0.15\textwidth}
        \centering
        \includegraphics[height=\linewidth]{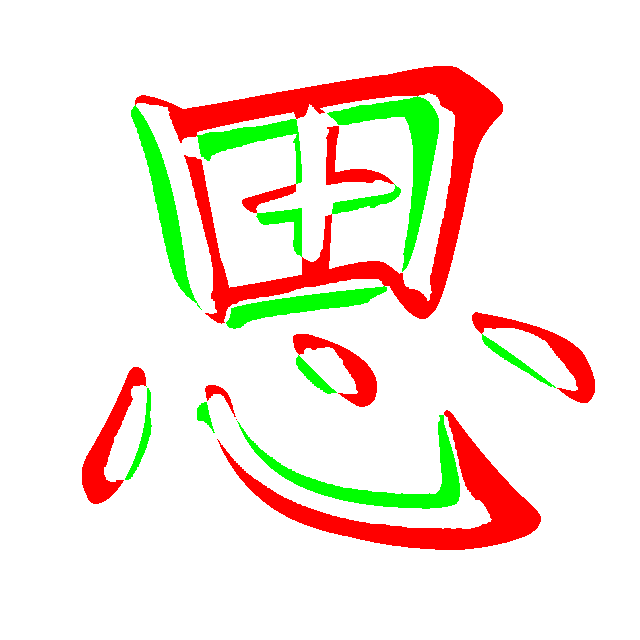}
        \caption{}
        \label{initial_dynamic2}
    \end{subfigure}
    \begin{subfigure}[b]{0.15\textwidth}
        \centering
        \includegraphics[height=\linewidth]{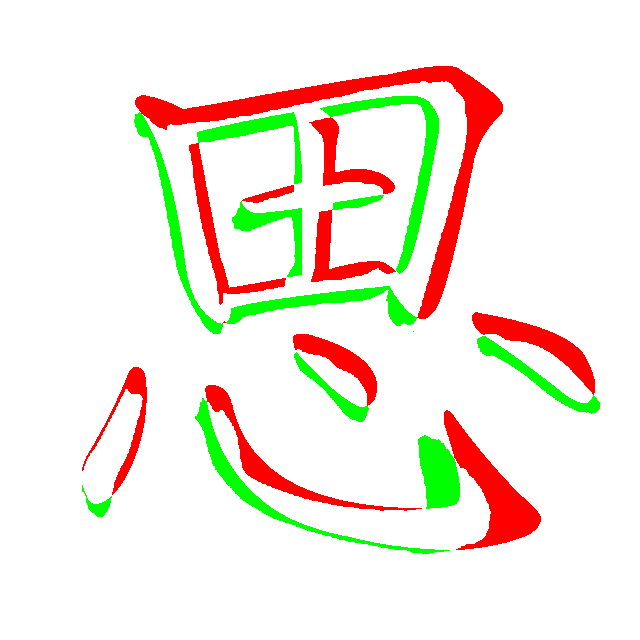}
        \caption{}
        \label{initial_dynamic3}
    \end{subfigure}
    \caption{\label{fig:comp-simple} Character `si', meaning `think'. (a) Written result following \textit{simple} brush optimization; (b) Written result following \textit{dynamic} brush optimization; (c) Written result from (a) overlaid with original SVG file; (d) Written result from (b) overlaid with original SVG file. The green and red pixels represent positive and negative differences respectively.}
\end{figure}

\begin{figure*}[t]
    \centering
    \includegraphics[width=0.99\linewidth]{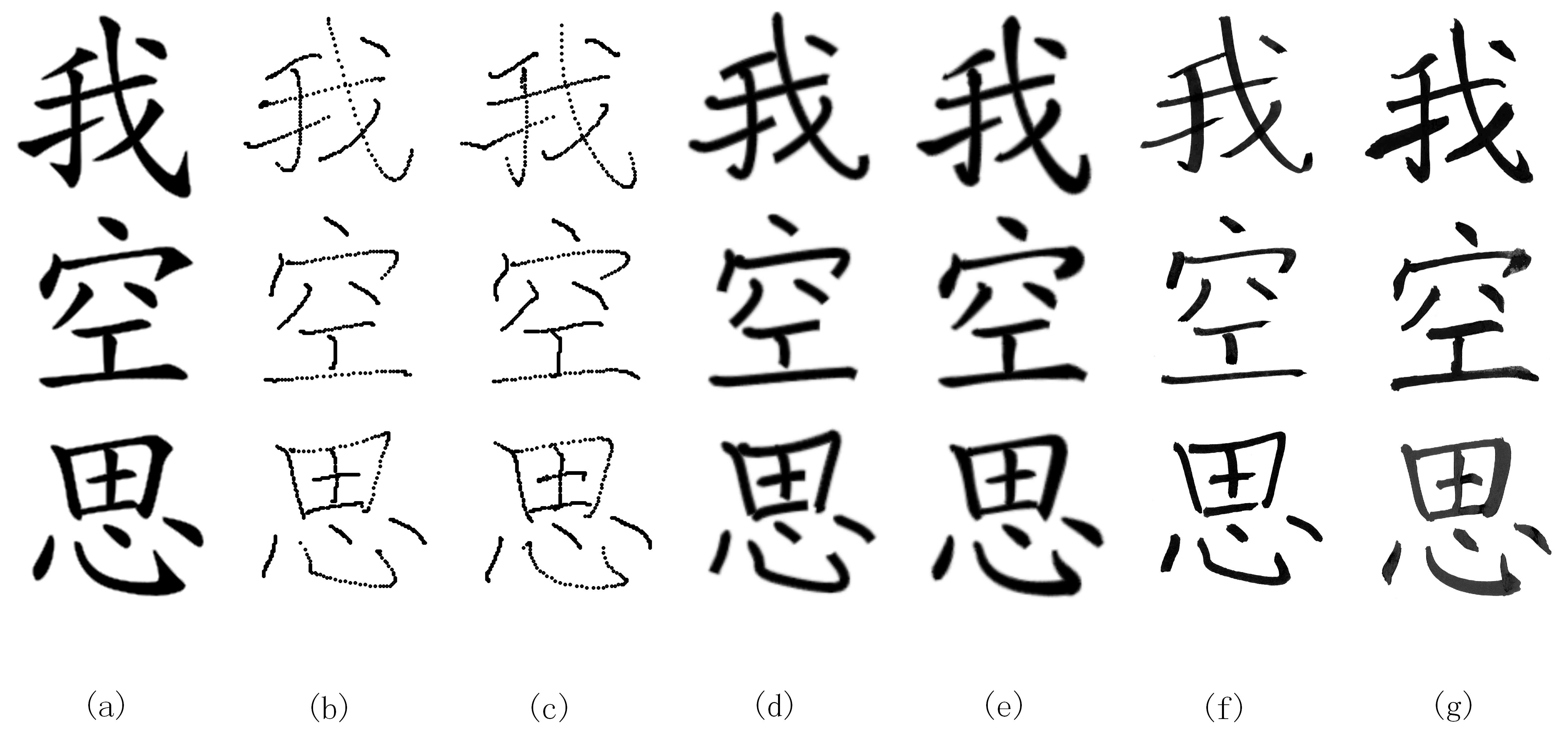}
    \caption{The optimization of different characters: from top down, `wo', `kong', and `si', meaning `me', `empty', and `think'. (a) The original character pictures from the database; (b) Initial trajectory estimates; (c) The trajectory obtained from optimization; (d) Simulated image drawn by the virtual brush using the initial trajectory; (e) Simulated image drawn by the virtual brush using an \textit{optimized} trajectory; (f) Written image following initial trajectories; (g) Written image following the \textit{optimized} trajectories.
    \label{fig:all_results}}
\end{figure*}

\begin{figure*}[t]
    \centering
    \includegraphics[width=0.88\linewidth]{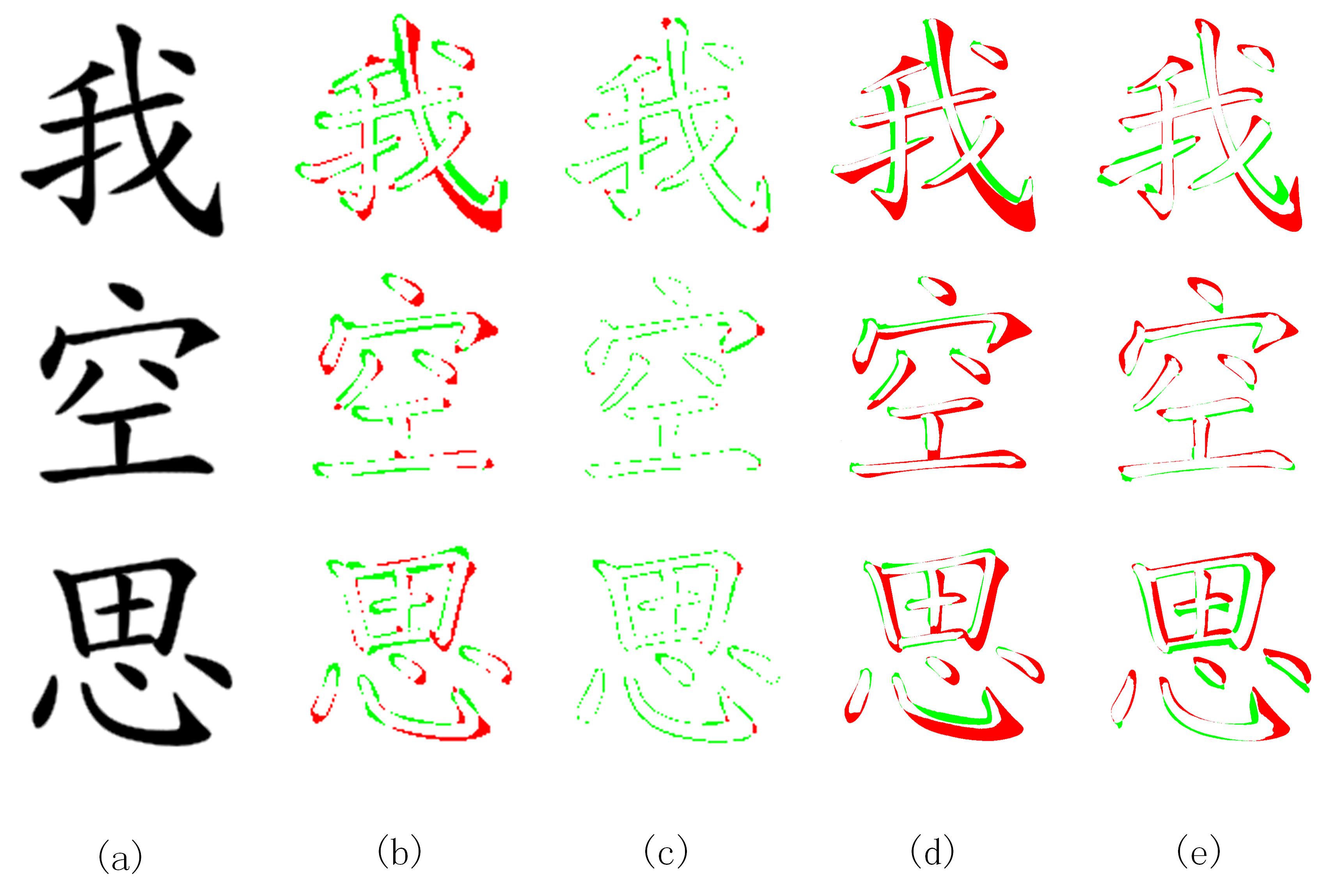}
    \caption{Error analysis for the three characters from Figure \ref{fig:all_results}. As before, green and red pixels represent positive and negative differences, respectively. (a) The original character pictures from the database; (b) Simulated image drawn by the dynamic virtual brush using the initial trajectory; (c) Simulated image drawn by the dynamic virtual brush using an \textit{optimized} trajectory; (d) Actual written image following initial trajectories; (e) Actual written image following the \textit{optimized} trajectories.
    \label{fig:differences}}
\end{figure*}

Below we present the results of our approach, including photographs of characters that have been drawn by a Franka robot, using inverse kinematics provided by
MoveIt!\cite{Coleman14arx_ReducingTB}. The characters are written using relatively slow end-effector velocity to prevent excessive jerk and vibrations.

Fig. \ref{fig:comp-simple} shows a comparison between the written results generated by the simple virtual brush model and by the dynamic virtual brush model. 
The simple virtual brush model is able to generate good touching and brush outlines for some simple strokes (like dot) because of the good linear fitting model in the Sim2Real section, but not all (like the long hook). However, it can be seen that the dynamic virtual brush shows obvious advantages in terms of preserving the relative positions of strokes, since it models the deformation of the real brush. Periodically restoring the brush to its initial state helps the simple brush to behave better, but it is still worse than the dynamic virtual brush model. 




In Fig.~\ref{fig:all_results}, we show more extensive results for three different characters.
Both the simulated and written images before and after optimization are shown for easy comparison. 
From the figure, we can see that the optimization achieves good performance for simulated images. However, because of a Sim2Real gap in the virtual brush model, the proposed method still has a way to go in terms of approaching the smoothness and definition of detail displayed in the reference images, rendered from the vector-based character database.

A detailed error analysis shows that the optimization achieves almost perfect results in simulation, but that there still is a considerable Sim2Real gap. Fig. \ref{fig:differences} shows the differences between reference images and both simulated and written characters, respectively before and after optimization. As in Figure \ref{fig:comp-simple}, green and red pixels represent positive and negative differences. Column (c) shows that the optimization can achieve near-perfect results, where the only errors remaining stem from anti-aliasing, as well as a few small errors near the stroke endings. However, we see larger errors when these optimized results are executed on the robot (column e), even though the characters themselves look visually pleasing.


\section{Discussion and Future Work}

In this paper, we presented a trajectory optimization method to make robots write calligraphy, searching for open-loop control trajectories to approximate a reference image.  The proposed dynamic virtual brush model yields good results in simulation and in turn produces reasonable open-loop control trajectories for a real robot. However, from the results it is clear that the Sim2Real gap has not been fully closed, and it is an open question whether a closed loop strategy will ever yield master-level calligraphy. Hence, in future work we plan to investigate both more predictable brush models to reduce the uncertainty, as well as the possibility of adding feedback control around optimized trajectories. 



\bibliographystyle{ieeetr}
\bibliography{painting} 

\end{document}